\documentclass[10pt]{article}

\usepackage{microtype}
\usepackage{graphicx}
\usepackage{subfigure}
\usepackage{booktabs} 

\usepackage{amsmath}

\usepackage{hyperref}

\usepackage[accepted]{icml2018}

\icmltitlerunning{Contextual Graph Markov Model}

\begin{document}

\twocolumn[
\icmltitle{Contextual Graph Markov Model: \\
           A Deep and Generative Approach to Graph Processing}

\icmlsetsymbol{equal}{*}

\begin{icmlauthorlist}
\icmlauthor{Davide Bacciu}{pi}
\icmlauthor{Federico Errica}{pi}
\icmlauthor{Alessio Micheli}{pi}
\end{icmlauthorlist}

\icmlaffiliation{pi}{Department of Computer Science, University of Pisa}

\icmlcorrespondingauthor{Davide Bacciu}{bacciu@di.unipi.it}
\icmlcorrespondingauthor{Federico Errica}{f.errica@protonmail.com}
\icmlcorrespondingauthor{Alessio Micheli}{micheli@di.unipi.it}

\icmlkeywords{machine learning, graphs, learning in structured domain, contextual models, deep learning, generative models}
\vskip 0.3in
]

\setlength{\abovedisplayskip}{5pt}
\setlength{\belowdisplayskip}{3pt}

\printAffiliationsAndNotice{\icmlEqualContribution}

\begin{abstract}
We introduce the Contextual Graph Markov Model, an approach combining ideas from generative models and neural networks for the processing of graph data. It founds on a constructive methodology to build a deep architecture comprising layers of probabilistic models that learn to encode the structured information in an incremental fashion. Context is diffused in an efficient and scalable way across the graph vertexes and edges. The resulting graph encoding is used in combination with discriminative models to address structure classification benchmarks. 
\end{abstract}

\newcommand{\quotes}[1]{``#1''}
\section{Introduction}
\label{sect:intro}
Learning in structured domains (SDs) deals with data of varying size and topology and amounts to identifying, synthesizing and embedding structural relationships into the model.
A naive approach, widely diffused in the past, would preprocess  the structure to obtain a fixed vectorial representation of hand-designed features, and feed it to standard learning models. In doing so, many of the relationships carrying useful information are definitively lost.
Recurrent and recursive models have been widely used to learn such encodings by imposing a topological order on directed acyclic structures (sequences, trees and DAGs) known as the \textit{causality} assumption which, however, restricts the classes of data that can be handled. In particular, the recursive definition of the state space cannot be applied to cyclic inputs without expensive diffusion mechanisms and constraints on the state transition function of the model.
The paper introduces a novel generative model for learning unsupervised encodings of cyclic structures, that borrows from previous SD processing works, 
namely the generative Bottom-Up Hidden Tree Markov Model \cite{bhtmm} and the constructive approach 
of Neural Network for Graphs \cite{nn4g}. We believe this to be the first scalable realization of a generative model for the processing of variable size graphs. Its probabilistic nature allows to 
model labeling uncertainty as well as 
to perform inference on graph vertexes and arcs. The unsupervised generative encoders can capture rich patterns in the data that do not depend on the problem at hand, allowing wide reuse of the encoding across different tasks and to eventually leverage unlabeled examples to boost accuracy in a semi-supervised setting.
In addition, the model is trained in an incremental fashion, which allows automatic construction of the architecture in supervised tasks. Information diffusion is achieved thanks to layering in a 
deep learning fashion. Computation within each layer is fully local, highly scalable and does not require iterative procedures or convergence conditions. Depth has a profound impact on context shaping since it allows each vertex to gather a wider picture of its surroundings, with a number of layers that is functional to the generative contextual encoding.
We will focus on classification tasks that require a transduction $\mathcal{T}$ from the domain of graphs to a discrete or continuous space. Such mapping has the advantage to be adaptive, in contrast to kernel functions which are fixed, and it is built by composition of an encoding function $\mathcal{T}_{enc}$ and an output function $\mathcal{T}_{out}$. The former takes as input a graph $\mathbf{g}$ and produces an hidden representation $z(\mathbf{g})$, while the latter takes such encoding as input and outputs the prediction. 
We will consider an output function implemented through a Support Vector Machine (SVM), taking in input the hidden representation $z(\mathbf{g})$ learned by the latent variables of the generative model.
The representational capability of our model, referred to as Contextual Graph Markov Model (CGMM), is tested on popular benchmarks for tree classification and on biochemical datasets where compounds are naturally represented as undirected graphs.


\section{Background}
\label{sect:back}
\begin{figure}[t]
\begin{center}
\centerline{\includegraphics[width=.5\columnwidth]{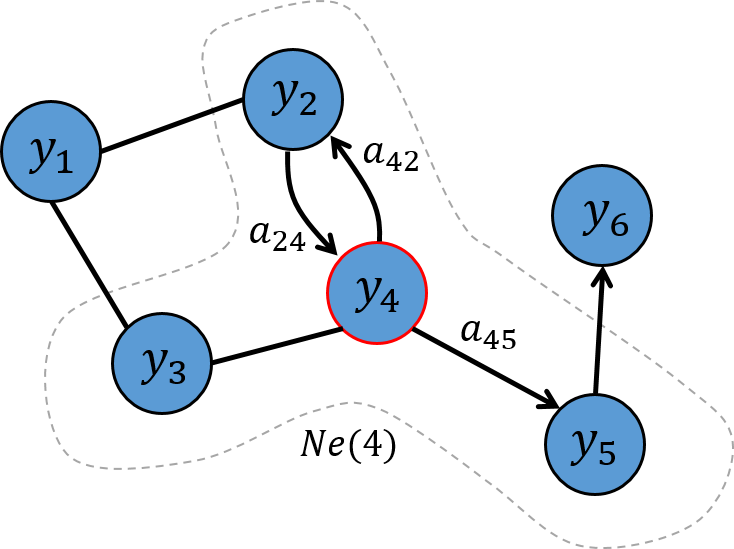}}
\caption{Example graph highlighting the notation used throughout the paper: $y_u$ denotes the label of a vertex $u$, $a_{uv}$ is the label of the edge connecting $u$ to $v$, while $Ne(u)$ is the set of neighbors of a vertex $u$. The figure highlights the neighborhood of vertex $4$ as a dashed area. Undirected edges denote the presence of two arcs oriented in different directions between the vertices and sharing the same label $a_{uv} = a_{vu}$. In general, a graph can include both directed and undirected edges. \label{fig:graph}}
\end{center}
\vskip -0.2in
\end{figure}

Let us first define the class of graphs we deal with in the rest of the paper together with the associated notation. We consider the problem of learning from a population of graphs structured data $\mathcal{G} = \{\mathbf{g}_1,\dots,\mathbf{g}_N\}$ where each sample $\mathbf{g}_n$ is a labeled graph with varying topology, number of nodes and edges with respect to the other samples. No restrictions are posed on the graph structure, in particular in terms of acyclicity. A graph $\mathbf{g} = (\mathcal{V}_g,\mathcal{E}_g,\mathcal{Y}_g,\mathcal{A}_g)$ is defined by a set of vertexes $\mathcal{V}_g$ and by a set of edges (also referred to as arcs) $\mathcal{E}_g$ between two vertexes. A directed arc ($u$,$v$) can be associated to a label $a_{uv}$ which, for the purpose of this paper, we assume to be taken from a finite alphabet $\{1,\dots,A\}$. Similarly, a vertex $u$ is associated to a label $y_u$ from a finite vocabulary $\{1,\dots,M\}$. When clear from the context, we omit the subscript $g$ to avoid cluttering. Figure \ref{fig:graph} shows an exemplar graph to clarify the notation: we use undirected edges, to ease the drawing, to denote the presence of two oriented arcs having the same label. In the remainder of the paper, we will use the concept of neighborhood of a vertex $u$, defined as $Ne(u) = \{ v \in \mathcal{V}_g | (v,u) \in \mathcal{E}_g\}$. In the following, we will consider an \textit{open} neighborhood of $u$ i.e. one that does not include $u$ itself. Undirected connections are modeled by two oriented arcs. Figure \ref{fig:graph} shows an example of the neighborhood for a target node $u = 4$, depicted as a dashed area.

Processing structured data requires to cater for samples that change size and connectivity, extracting from this variability patterns useful for predictive or explorative analyses. Several learning models have been dealing successfully with sequences and trees. When moving to more general and possibly cyclic topology, as in this work, things get considerably more cumbersome. Literature in this respect reports a number of works dealing with the problem of loops through 
simplifications and relaxations of the original problem.

Within the neural paradigm, two foundational approaches have been proposed, almost at the same time, to design neural networks for graphs. They are able to learn both the hidden representation (state encoding) and the output function for classification or regression tasks. 
By exploiting weight sharing techniques they also implement a stationarity assumption in order to efficiently treat variable-size structures.
The Graph Neural Network \cite{gnn} (GNN) is a supervised model using a direct extension of 
recursive models.
In an iterative fashion, at each step, information is exchanged between neurons associated to adjacent vertexes. Contractive state transition functions (obtained by weight constraints) are implemented by the neurons to ensure that the diffusion process can reach an equilibrium at each epoch. Hence GNN can process directed, undirected and cyclic graphs, at the cost of introducing a constraint on the weights values and a recursive dynamics for cycles that have drawbacks in terms of efficiency.
The Neural Network for Graphs \cite{nn4g} (NN4G), on the other hand, takes advantage of a feed-forward neural network construction whose layered 
architecture is determined during training e.g by a Cascade Correlation approach \cite{cascadecorrelation}. Cyclic structures are easily handled without recursive state definitions because computation is local to each vertex by exploiting frozen states, while the incremental construction allows to capture larger and larger contexts in a symmetric way (as formally proved in \cite{nn4g}). It follows that NN4G can be applied to any kind of graphs (as for GNN) despite the constraints on weights values being relaxed. NN4G also achieves a more efficient approach both by avoiding the need of a dynamical process for cycles and by exploiting an incremental approach to the task through progressive layers. 
In \cite{duvenaud2015convolutional} it has been proposed a hierarchical approach akin to NN4G and inspired by circular fingerprints in chemical structures. Differently from NN4G (and CGMM), the graph encoding is learned by end-to-end backpropagation on all layers, instead of incrementally. More recently, alternatives have been proposed extending Convolutional Neural Networks (CNN) to graphs: PATCHY-SAN \cite{cnn4g} has been the first to extend convolution to adjacent nodes in a graph by centering it on vertexes. When dealing with a graph dataset of varying topology, such an approach requires to determine a vertex ordering for each graph which is consistent across the entire dataset, as well as to impose an upper bound on the neighborhood size, limiting flexibility and context propagation. Two related approaches \cite{bresson16,kipf} extended CNN to graphs by exploiting a spectral analysis of the graph's Laplacian, which however limits applicability to a single graph of fixed size with undirected edges. Very recently \cite{Jure17}, it has appeared a preliminary work on extending graph convolutions to structures of varying size, using an information spreading mechanism very similar to that of \cite{nn4g}.

Kernels have been the most popular 
methodology for dealing with graph data in the last ten years. They are typically based on a relaxation of the original problem, counting matchings between simpler (computationally affordable) substructures/features in the original graphs, e.g. those generated by random walks from each 
vertex \cite{shortestpathkernelsongraphs,marginalizedkernels}. The solution proposed by \cite{treekernelnavarin} is to transform the graph into a multi-set of Directed Acyclic Graphs (DAGs) and to define 
similarity as a multi-set similarity score on the DAGs. The Fast-Subtree \cite{fastsubtreekernelsongraphs} kernel (FS) compares graphs by means of the Weisfeiler-Lehman (WL) test for graphs' isomorphism. This is based on an iterative algorithm which outputs a relabeled graph, where each vertex 
is the result of a multi-set compression. The FS kernel uses the relabeling from different WL iterations as structural features, fundamentally computing graph similarity in terms of the number of common strings in 
such relabeling. The use of 
these kernels on large datasets is limited by their computational costs. Further, they lack adaptivity, 
since the similarity function is defined by hand-engineered features rather than being learned from the data, which might limit the extent of their applicability.

The application of probabilistic models to graph data is largely 
constrained by the impact of cycles on the probabilistic relationship between 
random variables used to encode structural information.
 Cycles make the inferential and training procedures too complex for practical 
 use.In this respect, much of the contributions have been dealing with acyclic structures such as trees, typically through recursive extensions of the Hidden Markov Model 
\cite{frasconi1998general, htmm, bhtmm}. These probabilistic models learn a distribution $P(\mathbf{x})$ over a tree $\mathbf{x}$ assuming nodes are generated by hidden states, with a transition probability that depends on the node's children, siblings or parents.
These (typically unsupervised) generative models can be used to solve supervised tasks, e.g. by defining generative kernels on top of the learned probabilistic process as in \cite{genkerntreestructureddata}.

The model proposed in this paper introduces the first scalable generative model for graphs of varying size, by building on the context propagation scheme of NN4G. It uses hidden variables to encode structural information using diffusion from neighboring nodes. In order not to incur in problems with cycles, 
context propagation is achieved thanks to the frozen representation of the neighbors' hidden states at preceding generative layers. 

\section{Contextual Graph Markov Model}
\label{sect:cgmm}
In the following, we introduce the formalization of the Contextual Graph Markov Model (CGMM)\footnote{\url{https://github.com/diningphil/CGMM}}, providing a brief summary of the main procedures required to run the model, i.e. parameter learning, inference of the vertex encoding and incremental layering policies.

\subsection{The model} \label{sect:model}
CGMM is a probabilistic model for graphs that learns to encode structural information, adopting a modular approach that exploits both a stack of 
base modules and a layer-wise pooling strategy to increase 
discriminative efficacy along the lines of \cite{cascadecorrelation}. As anticipated, the CGMM architecture borrows from NN4G, whereas the realization of 
each layer is strongly influenced by the recursive probabilistic models for trees, such as the 
Bottom-up Hidden Tree Markov Model (BHTMM) \cite{bhtmm}.

\begin{figure}[t]
\begin{center}
\centerline{\includegraphics[width=\columnwidth]{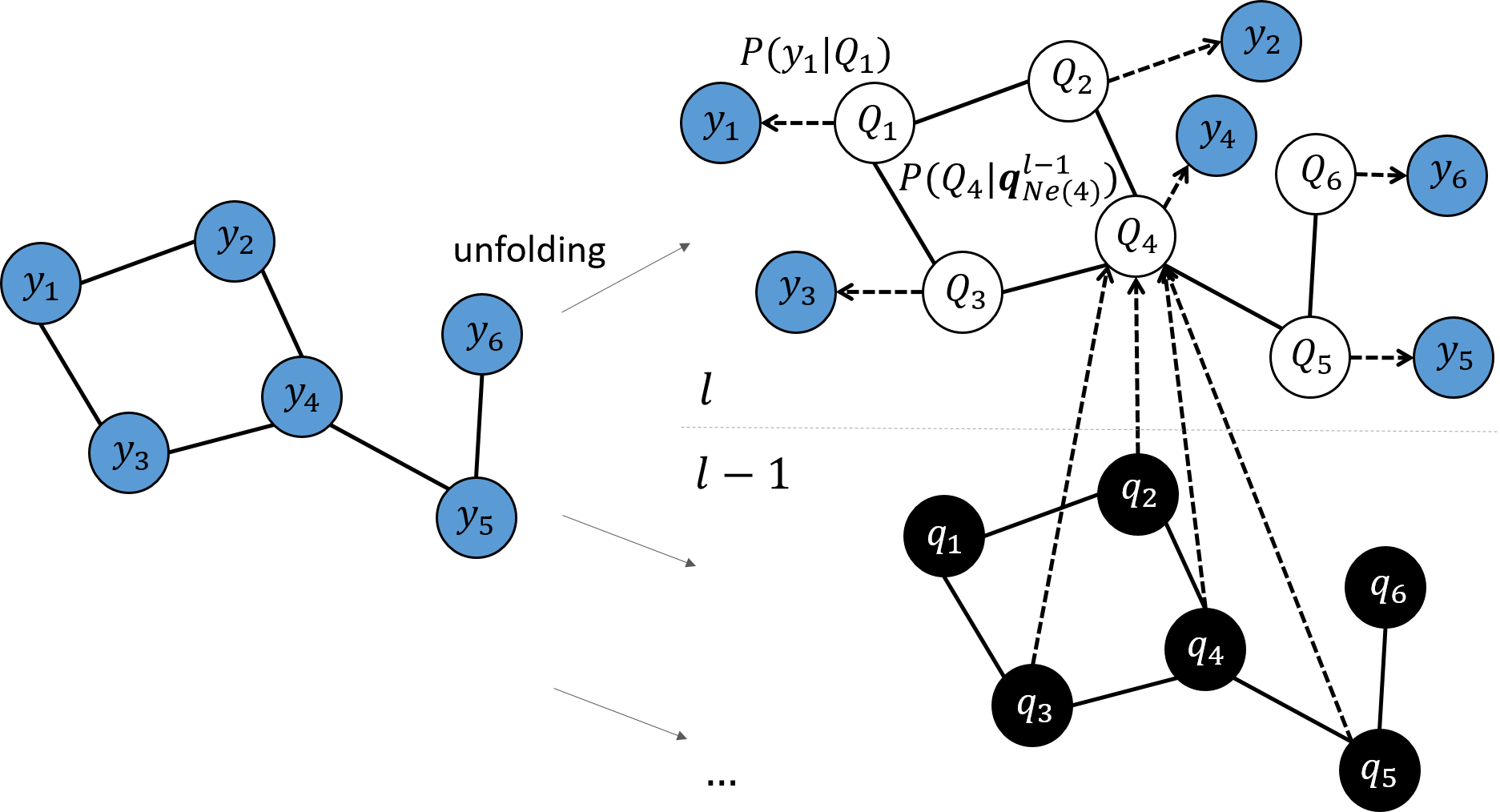}}
\caption{Representation as a graphical model of the unfolding of the $l$-th layer of the CGMM on the structure of the graph on the left. As usual, empty nodes correspond to unobserved variables, while full nodes denote observed vertex labels (shaded) and state assignments at previous level (black nodes). Thick undirected edges denote arcs while dashed arrows represent probabilistic causation relationships. \label{fig:unfold}}
\end{center}
\vskip -0.2in
\end{figure}

%
The base CGMM component models a graph using discrete hidden state variables which encode information regarding vertexes and their context. Each vertex $u$ is thus associated with a multinomial random variable $Q_u$ with values on the finite alphabet $\{1,\dots,C\}$, where $C$ is the hidden state size. Like in hidden Markov models, the hidden state assignment for a node $u$ determines the emission of the observed label $y_u$ (again assumed discrete) through the emission distribution $P(y_u| Q_u)$. Differently from a standard hidden Markov model, however, the current hidden state assignment, i.e. at layer $l$ of the architecture, is not determined by a transition from another set of hidden states at the same layer $l$. 
Instead, a vertex state $Q_u$ at layer $l$ is determined by the 
frozen states' assignments of the neighboring vertexes $Ne(u)$ at previous layers $l' < l$. 
This is a key difference as such assignments can be considered fully observable at level $l$, thus avoiding mutual causal dependencies between the layers and preventing to get stuck in indefinite inference loops due to cycles.
Figure \ref{fig:unfold} shows a representation of the approach as a graphical model, focusing on a target layer $l$. Hidden states at current level are represented by capital $Q_u$ terms, which are empty nodes of the graphical model, since they are unobserved. Vertex labels are observed nodes $y_u$ and the probabilistic model unfolds on the structure of the graph like a hidden Markov model unfolds on the structure of a sequence. Figure  \ref{fig:unfold} also depicts how the hidden state for node $u=4$ is determined through the distribution $P(Q_u|\mathbf{q}_{Ne(u)}^{l-1})$ based on the state of its neighbors (and possibly himself) at the previous level $l-1$. Such states are determined by another probabilistic model, trained previously and then frozen. The resulting state assignments are marked with lowercase $q_u$ terms and the corresponding nodes are black to denote the fact that they have to be considered observed.

\begin{figure}[t]
\begin{center}
\centerline{\includegraphics[width=.7 \columnwidth]{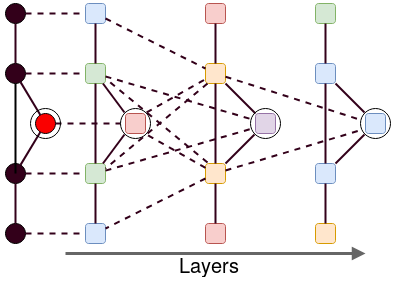}}
\caption{Unfolding of the hidden state variables for a cyclic structure (on the left).
In this diagram, solid lines represents edges while dashed lines represent the flow of contextual information. At layer 1 vertexes are unaware of any information concerning surrounding structures. If we focus on the red vertex we see how increasing the depth leads to states (rounded squares) that encode a larger portion of the graph. Perhaps more importantly, the spreading of information is symmetric thanks to relaxation of the causality assumption. This cannot be achieved by a recurrent or recursive model without introducing a cyclic state definition. \label{fig:flow}}
\end{center}
\vskip -0.3in
\end{figure}

The scheme depicted in Figure \ref{fig:unfold} is iterated across a number of layers, each independently trained with respect to the previous ones but using the knowledge extracted by them under the form of hidden state labels. Figure \ref{fig:flow} shows the unrolling of the CGMM on a four-layers structure. The process starts at level one where hidden states are assigned without taking any context into consideration, except for the vertex label. At next iteration, vertexes have access to information concerning their direct neighbors. At level $3$ they start receiving information from the neighbors of their neighbors. The iteration of this process allows an effective context propagation from each node of the graph (provided that a sufficient number of layers is used).
The propagation of context in this architecture is symmetric with no need for causal dependencies between hidden states. As a consequence, CGMM can efficiently process graphs of any dimension without assumptions on topological properties, in contrast with models that need limiting the size of the vertex neighborhood or add padding to work with a fixed-size representation.

Each layer of the CGMM is implemented using different instantiations of the same probabilistic model, for which we provide the likelihood function in the following. Let us start by defining $L^{-1}(l)$ as the set of layers $l'$ that precede current layer $l$. We let $\mathbf{\hat{q}^{L^{-1}(l)}_{Ne(u)}}$ denote the set of neighboring hidden states for vertex $u$, provided by previous layers $l' \in L^{-1}(l)$. We can then define the likelihood of a model at level $l$ as
\begin{equation}
\label{eq_likelihood}
\mathcal{L}(\theta|\mathcal{G}) = \prod_{\mathbf{g} \in \mathcal{G}} \prod_{u=1}^{\mathcal{V}_g} \sum_{i=1}^{C} P(y_u | Q_u = i) P(Q_u = i | \mathbf{\hat{q}}^{L^{-1}(l)}_{Ne(u)}).
\end{equation}
Please recall that, when clear from the context, we will abstract from the indexing term $g$ to ease notation. Equation (\ref{eq_likelihood}) assumes that vertexes are independent from each other, which makes training of the model efficient and scalable. Structural information is indirectly conveyed through conditioning on the state assignment from previous layers. The complexity of the rightmost term can become infeasible due to the (potentially) combinatorial number of states in the conditioning part. This can be made tractable by resorting to a Switching Parent (SP) approximation such as the one in \cite{bhtmm}, so as to control complexity of children-to-parent transitions. Here, we introduce a SP variable $L_u$ whose probability $P(L_u = l')$ denotes the weight of level $l'$ in determining the state transition, and another variable $S_u$ whose distribution $P^{l'}(S_u = a)$ controls the weight that an arc with label $a$ should be given when the layer considered is $l'$. The resulting likelihood is
\begin{align}
\mathcal{L} = & \prod_{\mathbf{g} \in \mathcal{G}} \prod_{u=1}^{\mathcal{V}_g} \sum_{i=1}^{C} P(y_u | Q_u = i) \sum_{l' \in L^{-1}(l)} P(L_u = l') \nonumber \\
& \sum_{a=1}^A P^{l'}(S_u = a)P^{l',a}(Q_u = i |\mathbf{\hat{q}}^\mathbf{l',a}_{Ne(u)})
\label{eq:decomposition}
\end{align}
where we have introduced $\mathbf{\hat{q}}^{l',a}_{Ne(u)}$ to denote the states' assignments at level $l'$ of all neighbors of $u$ having label $a$ on their arc. 
Such a parameterization allows to differentiate vertexes based on their hidden state, the label on the connecting edge as well as the depth where the contextual information is generated. The hidden state distribution can be further decomposed as
%
%
\begin{align*}
P^{l',a}(Q_u = i |\mathbf{\hat{q}}^\mathbf{l',a}_{Ne(u)}) = \frac{\sum_{v \in Ne^{l',a}(u)}P^{l',a}(Q_u = i | q_v)}{|Ne^{l',a}(u)|}\\
\end{align*}
assuming that all vertexes in $Ne^{l',a}(u)$ which share the arc label $a$ contribute equally (for each level $l'$). Clearly, alternative decompositions of this distribution can be devised,  but for the sake of this paper we stick to the most simple and intuitive.

Summarizing, the model parameters for a level $l$ are the emission distribution $P(y | Q)$, the arc-and-level-dependent state distribution $P^{l',a}(Q_u  | q_v)$, plus the two weighting terms $P(L = l)$ and $P^l(S = a)$ (all multinomials). 
%
%
All are shared between vertexes allowing the model to easily scale to graphs of any size. The first CGMM layer has no previous hidden states as inputs, requiring to fit only prior and emission distributions.
Model complexity is dominated by $\mathcal{O}(\mathcal{V}_g*|L^{-1}(l)|*A*C^2)$ and can be regulated by tweaking the number of preceding layers $L^{-1}$ and hidden states $C$. In its current implementation CGMM can process $20-22$ vertices per $1$ms.
%
%

\subsection{Training}
Learning of a CGMM is achieved by training independently the layers composing the model. This can be performed by maximization of the likelihood in (\ref{eq:decomposition}) using an Expectation-Maximization approach.
Given the space limitations, we consider vertex labels to be discrete and unidimensional, but an extension to multidimensional and continuous labels is straightforward.

Formally, we introduce into the likelihood the indicator variables $z_{uilaj}$, that are one when vertex $u$ is in state $i$ while its neighbors with arc label $a$ are in state $j$ at level $l$, and 
zero otherwise. By these means we can rearrange the terms in Equation (\ref{eq:decomposition}) to comprise only multiplications, which are later transformed into summations through the application of a logarithm. When applying the E-Step on such a function, we are seeking the posterior of the indicator variables, which can be shown to be equivalent to
\begin{align}
\begin{split}
  E[z_{uilaj} | \mathbf{g,\hat{q}}] & = P(Q_u = i, L_u = l, S_u = a, q = j | \mathbf{g,\hat{q}})\\
  = & \frac{1}{Z} P(y_u | Q_u = i)P^{l,a}(Q_u = i | q = j)\\
  & \ \ \ \ \ \times P(L_u = l)P^{l}(S_u = a)
\end{split}
\label{eq:post}
\end{align}
where $Z$ is a normalization term which is straightforward to compute. Such posterior probabilities are then used to update the model parameters at the M-Step
\begin{align*}
& P(y = k | Q = i) = \frac{1}{Z}\sum_{\substack{\mathbf{g} \in \mathcal{G} \\ u \in \mathcal{V}_g}} \delta(y_u, k) E[z_{ui}|\mathbf{\mathbf{g},\hat{q}}]\\
& P(L = l) = \frac{1}{Z}\sum_{\substack{\mathbf{g} \in \mathcal{G} \\ u \in \mathcal{V}_g \\ i \in \{1,\dots,C\}}} E[z_{uil}|\mathbf{\mathbf{g},\hat{q}}]\\
\end{align*}
\begin{align*}
& P^{l}(S = a) = \frac{1}{Z}\sum_{\substack{\mathbf{g} \in \mathcal{G} \\ u \in \mathcal{V}_g \\ i \in \{1,\dots,C\}}} E[z_{uila}|\mathbf{\mathbf{g},\hat{q}}]\\
& P^{l,a}(Q = i | q = j) = \frac{1}{Z}\sum_{\substack{\mathbf{g} \in \mathcal{G} \\ u \in \mathcal{V}_g}} \Big( E[z_{uilaj}|\mathbf{\mathbf{g},\hat{q}}]\Big)
\end{align*}
where $\delta$ is the Kronecker delta. The different posteriors reported above can be obtained by straightforward marginalization of $E[z_{uilaj} | \mathbf{g,\hat{q}}]$.

\subsection{Inference}
Inference serves to determine the most likely hidden state assignment given the observed data (at each level of the CGMM model). Specifically, for the first layer, we determine state assignment as
\begin{equation}\label{eq:infLev0}
\max_i P(Q_u = i | \mathbf{g}) = \max_i P(y_u | Q_u = i) P(Q_u = i)
\end{equation}
while for $l>1$ this is computed as
%
%
\begin{align}
\label{eq:infLevNext}
 & \max_i P(Q_u = i | \mathbf{g},\mathbf{\hat{q}}) = \max_i \Bigg\{ P(y_u | Q_u = i) \times \nonumber \\
 & \sum_{l'\in L^{-1}(l)} P(L = l') \sum_{a=1}^A P^{l'}(S = a) \times \nonumber \\
 & \sum_{v \in Ne^{l',a}(u)}\frac{P^{l',a}(Q_u = i | q_v)}{|Ne^{l',a}(u)|} \Bigg\} 
\end{align}
The latent state assignment is fundamental to propagate the context to the higher layers of the CGMM. At the same time, it can be used to obtain a fixed-size vectorial encoding of the graph in terms of its hidden states occurrences, as discussed in the next section.

\subsection{CGMM supervision, layering and pooling} \label{sect:super}
\begin{figure}[t]
\begin{center}
\centerline{\includegraphics[width=\columnwidth]{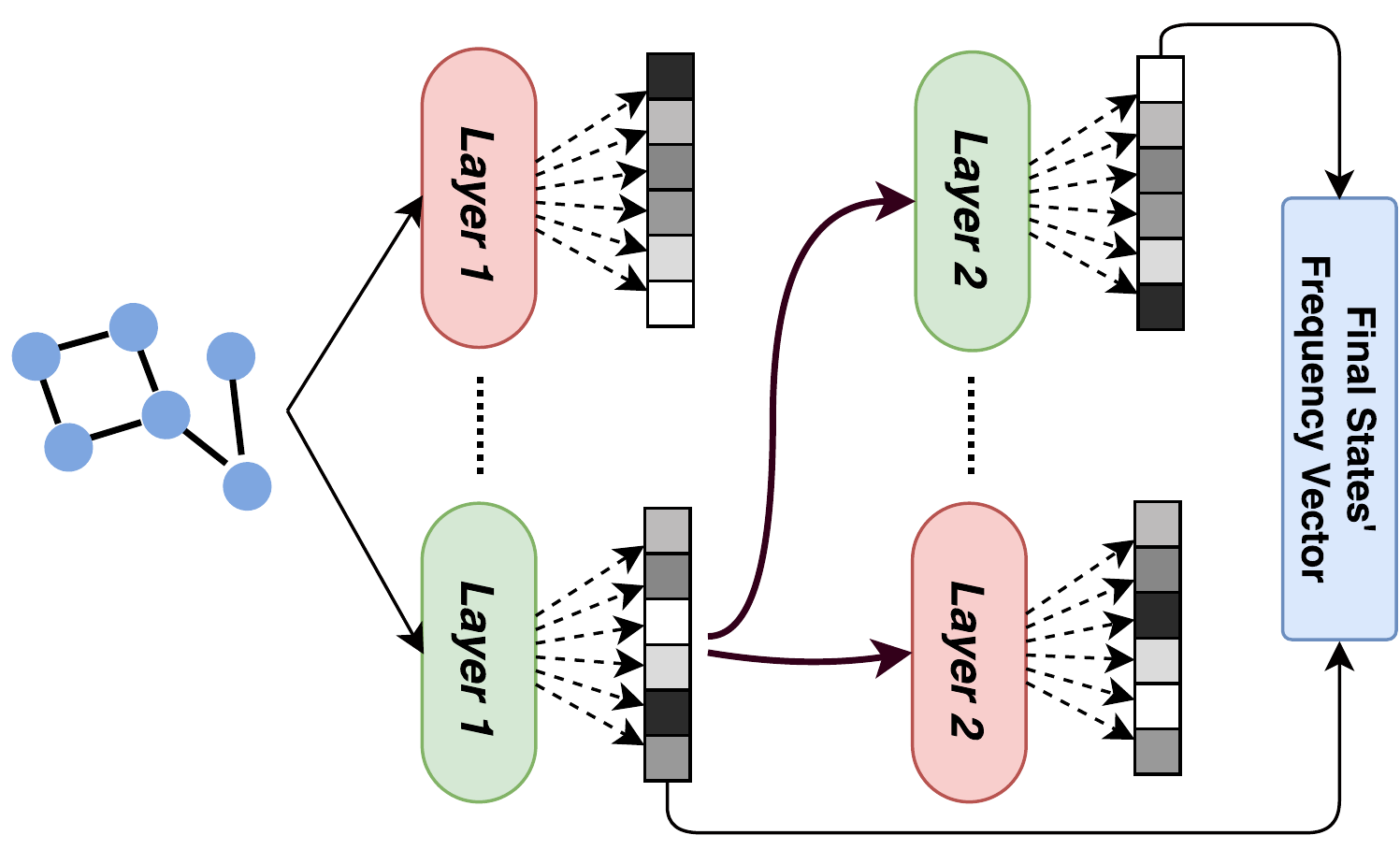}}
\caption{An overview of the training process for an architecture of final depth 2. At each layer a pool of CGMMs is trained and one of them is selected (shown in green) according to a supervised criterion. The second layer exploits the most likely hidden states inferred. States' frequency vectors computed at each layer can be concatenated to become the structure's fingerprint. \label{fig:overview}}
\end{center}
\vskip -0.3in
\end{figure}

The process described in previous sections serves to learn a probabilistic unsupervised encoder of graph structured information. In order to apply the CGMM approach in a supervised setting we need to define 
a mean to exploit the knowledge captured in its hidden state space. Following what has been proposed in \cite{generativemultisetkernel}, we can define an encoding of the graph 
as a vector of state frequency counts for each layer, which are then concatenated into a fixed-size vector 
summarizing contributions from all layers. Figure \ref{fig:overview} shows an high level view of such a process.  The encoding obtained from the unsupervised CGMM can then be used as input to a standard classifier/regressor for performing supervised tasks. In this paper, we will use a SVM to test the CGMM encoding in graph classification tasks.
Another question that needs answering to provide a fully functional and general implementation of the CGMM is: how do we determine the correct number of layers? Simple answer is, we rely on model selection, allowing it to decide how many layers are needed to achieve the best generalization performance. The procedure implemented in our CGMM is straightforward: after the addition of a new layer, this is first trained using the EM equations described 
above. Then a supervised model is trained and tested using the current graph encoding as input. If the newly generated layer has a positive effect on the performance, then we try adding another layer, otherwise we stop. 

A last feature that we have bundled in our implementation of the model is the use of pooling, motivated by the considerations in \cite{cascadecorrelation}: as shown in Figure \ref{fig:overview}, at each layer we do not train a single probabilistic model. Rather we train different randomly initialized models, testing their respective encodings as described before. Then, for each level, we pick the best performing model of the pool and use 
it as input for the next layer of the architecture.

\section{Experiments}
\label{sect:experiments}
The section provides an empirical assessment of 
CGMM's ability in extracting meaningful structural patterns from the data. Since 
our model is deeply rooted in hidden Markov models for trees, we first test it on tree structured data classification, confronting its performance with that of state-of-the-art probabilistic models and kernels for trees. Then we extend the analysis to more general structures testing CGMM on graph classification tasks.

\subsection{Tree Classification Tasks}
Previous works try to tackle tree classification by recursively unfolding structures in a top-down or bottom-up \cite{htmm,bhtmm,iobhtmm,bidirectional} fashion, i.e. from the root to the leaves or from the frontier to the root node.  The \textit{INEX2005} and \textit{INEX2006} \cite{inex} are intensively studied benchmarks in this context. They concern the classification of XML formatted documents from two IEEE \textit{structure only corpora}. These datasets are characterized by a large number of unbalanced classes and discrete node labels. Such trees are generally shallow and with large out-degree. Table \ref{tab:inex} summarizes the characteristics of the two datasets.
\begin{table}[tb]
\caption{Details of the \textit{INEX2005} and \textit{INEX2006} datasets. The total number of nodes is, respectively, 124,359 and 218,537.}
\label{tab:inex}
\vskip 0.15in
\begin{center}
\begin{small}
\begin{sc}
\begin{tabular}{lcccr}
\toprule
Data set & Size & Classes & Degree & Labels \\
\midrule
\textit{INEX2005} & 9361 & 11 & 32 & 366 \\
\textit{INEX2006} & 12107 & 18 & 66 & 65 \\
\bottomrule
\end{tabular}
\end{sc}
\end{small}
\end{center}
\vskip -0.1in
\end{table}

Train and test splits are defined by the benchmark and comprise about $50\%$ of the data each. Model selection decisions have been taken using an hold-out validation set of $20\%$ of the training data. In these tasks, we have considered a CGMM without pooling and with a number of layers ranging from $1$ to $4$, which are reasonable choices considering that the INEX structures are, on average, quite shallow. Rather than aiming at identifying the best number of layers, in this experiment we will concentrate on assessing the effect of layering on 
information diffusion. CGMM hidden state size $C$ has been chosen in $\{20,40\}$, while for context propagation we have considered only the influence from the immediate predecessor of the current layer. The undirected tree edges have been transformed into two directed edges, one incoming and one outgoing, to account for context propagation both in an upwards and downwards direction (as occurs in the Hidden Tree Markov Model thanks to the upwards-downwards inference). Edges have been labelled with the relative position in the children subtree of the parent node.

Tree classification has been performed by placing a SVM on 
top of the CGMM hidden state encodings as described in Section \ref{sect:super}. To compute similarity between 
these representations we have considered both a standard RBF kernel as well as a Jaccard kernel, such as 
the one introduced in \cite{genkerntreestructureddata} in its unigram or bigram variant. SVM hyperparameters $C_{svm}$ and $\gamma_{svm}$ have been selected from a limited set in $\{5,50,100\}$, as their choice had little impact on the model performance. The performance of CGMM is reported in Table \ref{tab:inexresults}, where 
it is compared to that of state of the art tree classification approaches. These include both the related input-driven generative model for trees (IOBHTMM) \cite{iobhtmm}, the syntactic PT kernel \cite{ptkernel}, the generative tree kernels IOBHTMM-J, AM-GTM and Fisher \cite{generativemultisetkernel} as well as the bi-directional BDIO-J kernel \cite{bidirectional}. For the INEX 2006 data we also provide results for TreeESN \cite{treeEsn}, a recursive neural network model based on the reservoir computing paradigm.
\begin{table}[t]
\caption{Test accuracies for \textit{INEX2005} and \textit{INEX2006}.}
\label{tab:inexresults}
\vskip 0.15in
\begin{center}
\begin{small}
\begin{sc}
\begin{tabular}{lcc}
\toprule
\textbf{Model} & \textbf{INEX2005} & \textbf{INEX2006} \\
\midrule
\textit{IOBHTMM} & 90.17 (3.13) & 27.61 (1.34)\\
\textit{TreeESN} & - & 42.62\\
\textit{PT} & \textbf{97.04} & 41.13\\
\textit{Fisher}  & 96.82 (0.1) & 38.47 (0.8)\\
\textit{BHTMM-J} & 96.12 (0.1) & 45.06 (0.2) \\
\textit{BDIO-J}  & 95.24 (0.17) & \textbf{45.19}(0.12)\\
\textit{CGMM-RBF}  & 96.60 (0.05) & 45.1 (0.2)\\
\textit{CGMM-Jaccard} & 96.73 (0.06) & 43.18 (0.3)\\ \\
\bottomrule
\end{tabular}
\end{sc}
\end{small}
\end{center}
\vskip -0.in
\end{table}

CGMM shows a competitive performance on both datasets, matching the results of the two best models in literature.
It is worth to note that CGMM relaxes the causality assumptions that in the BHTMM-based models are fundamental to capture the relationships between substructures composing the tree. Notwithstanding such a simplification, the context propagation mechanism 
allows CGMM to capture and summarize sufficient structural patterns to be competitive with approaches specialized in treating trees, which often result in high computational complexities such as for the Partial Tree (PT) \cite{ptkernel}.

\begin{figure}[t]
\begin{center}
\centerline{\includegraphics[width=\columnwidth]{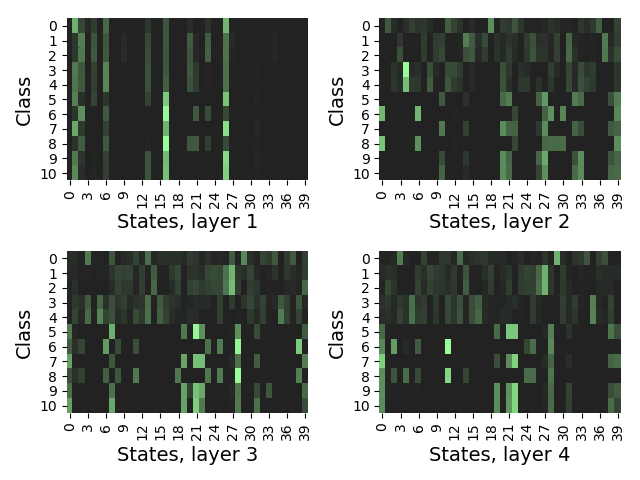}}
\caption{Average fingerprints for each target class of \textit{INEX2005}. The model has been trained with C=40. Lighter areas of the plot denote more \quotes{popular} states. \label{fig:inexavgfingerprints}}
\end{center}
\vskip -0.2in
\end{figure}

CGMM allows to gain an insight on the organization of the state space, by inspecting how the 
hidden states' activations behave at different layers and for different tree classes. Figure \ref{fig:inexavgfingerprints} 
presents an example of such hidden state fingerprint for the INEX 2005 classes, obtained by averaging hidden state activations of the composing trees. One can clearly appreciate the effect of context diffusion between first and second layer, where a certain number of states seem to become tuned to class-specific structural patterns. Although the use of more layer has a positive effect on accuracy, the differences between layers fingerprints fade for $l>2$. This is not surprising, considering the shallow nature of INEX trees
 for which a couple of layers is sufficient to fully propagate the context.

\subsection{Graph Classification Tasks}
The second part of the experimental analysis shows how CGMM 
effectively processes a more general class of structures than trees. To this end, we consider a set of standard graph classification benchmarks from the biochemical domain, that are MUTAG \cite{mutag}, CPDB \cite{cpdb} and AIDS \cite{st}. All benchmarks are binary classification tasks, where graph vertexes are labeled with discrete values representing atomic symbols. Edges are undirected and their label encodes the type of atomic bond: Table \ref{tab:chemicaldatasets} summarizes the statistics of the datasets. Again, we have tested the CGMM model with varying number of levels (from $1$ to $8$) and considering state information only from the direct predecessor layer. Both RBF and Jaccard kernels have been tested, as well as two different fingerprint construction strategies: one using the encoding produced by the last layer; the other using the concatenation of the encodings produced by all 
layers. The hidden states size has been chosen in $\{20,40\}$. A pooling strategy has been used 
with pool size set to $10$.
\begin{table}[t]
\caption{Statistics of the graphs in the biochemical datasets: the \quotes{Pos} column reports the percentage of members of the positive class, while nodes and edge values are averaged.}
\label{tab:chemicaldatasets}
\vskip 0.1in
\begin{center}
\begin{small}
\begin{sc}
\begin{tabular}{lcccr}
\toprule
Data set & Size & Pos(\%) & Atoms & Edges \\
\midrule
MUTAG & 188 & 66.48 & 45.1 & 47.1\\
CPDB & 684 & 49.85 & 14.1  & 14.6\\
AIDS & 1503 & 28.7 & 58.9 & 61.4\\
\bottomrule
\end{tabular}
\end{sc}
\end{small}
\end{center}
\vskip -0.1in
\end{table}

Experimental assessment is performed according to two model selection schemes, following 
the approaches used in literature. One assesses performance using a single cross-validation (CV) on the standard 10-fold splits provided for the dataset (i.e. performance is reported for the best model identified on the 10-fold validation error). The second approach, less used in literature but more robust, uses a nested CV, where a 5-fold CV is applied to each training fold of the outer 10-fold: in this case, model selection decisions are taken based on the internal 5-fold CV performance. Tables \ref{tab:chemicalmodelselection} reports the classification accuracy for the two assessment schemes: note that the set of models with which we confront varies in function of the method used in the respective papers.
To provide a comparison, we quote results from related approaches:
FS kernel \cite{fastsubtreekernelsongraphs}, WL-DDK kernels \cite{weisfeiler_navarin},
ODD-ST kernel \cite{treekernelnavarin}, Marginalised Graph Kernel (MGK) \cite{marginalizedkernels}, Correlated Pattern Mining (CPM) \cite{cpm}
gBoost \cite{gboost}, Convolutional Neural Network for Graphs (PATCHY-SAN) \cite{cnn4g}. We also provide results obtained by applying tree kernels to all the spanning trees generated 
from each graph vertex, using generative tree kernels from \cite{generativemultisetkernel} as well as syntactic tree kernels, such as ST \cite{st} and SST \cite{sst}.
The results for the MUTAG dataset are provided only for the single 10-fold CV as no model in literature appears to use it with a nested CV assessment (CGMM scores an accuracy of $85.3\%$ when assessed with a nested CV on MUTAG).
\begin{table}[t]
\caption{Performances (classification accuracy, in $\%$) for model selection and risk assessment.}
\vskip 0.05in
\begin{center}
\begin{small}
\begin{sc}
\begin{tabular}{lccr}
\toprule
 & CPDB & AIDS & MUTAG\\
\midrule
\textit{10-FOLD CV}\\
\textit{CPM} & 76.0 & 83.2 & 80.8\\
\textit{MGK} & 76.5 & 76.2 & \textit{-}\\
\textit{gBoost} & 78.8 & 80.2 & 85.2\\
\textit{Fast Subtree} & 76.3 & 79.1 & 89.3\\
\textit{$ODD-ST_h$} & 80.4 & 83.5 & 87.8\\
\textit{PATCHY-SAN} & \textit{-} & \textit{-} & \textbf{92.63} (4.21)\\
\textit{CGMM} & \textbf{81.04} (4.00) & \textbf{84.16} (2.31) & 91.18 (6.02)\\
& & &\\
\textit{NESTED CV}\\
\textit{IOBHTMM-J} & 69.03 (3.35) & 79.17 (3.46)\\
\textit{AM-GTM} & 75.44 (3.74) & 81.33 (3.89)\\
\textit{Fisher} & 68.87 (3.41) & 76.65 (3.45)\\
\textit{ST} & 75.29 (1.64) & 82.00 (2.00) \\
\textit{SST} & 76.59 (2.16) & 80.17 (1.53) \\
\textit{Fast Subtree} & 73.22 (0.78) & 75.61 (1.00)\\
\textit{$ODD-ST_h$} & 76.44 (0.62) & 81.51 (0.74)\\
\textit{$WL_{NS-DDK}$} & 77.03 (1.18) & 82.8 (0.66)\\
\textit{$WL_{DDK}$} & 76.52 (1.16) & 82.93 (0.71)\\
\textit{CGMM} & \textbf{78.06} (6.47) & \textbf{83.15} (2.17)\\
\bottomrule
\end{tabular}
\end{sc}
\end{small}
\end{center}
\vskip -0.2in
\label{tab:chemicalmodelselection}
\end{table}

Results highlight that CGMM obtains the highest accuracies on both CPDB and AIDS using assessment schemes compared to a number of state-of-the-art graph kernels of various nature (syntactical, convolutional, generative). The standard deviation of CGMM is relatively high on the CPDB dataset for what looks like an unfortunate split of the folds (only 2 folds out of 10 have accuracies considerably higher and lower than the average, hence the high variance).  Results on MUTAG are encouraging as well, basically matching the state of the art of the PATCHY-SAN neural model.

\begin{figure}[t]
\begin{center}
\centerline{\includegraphics[width=1\columnwidth]{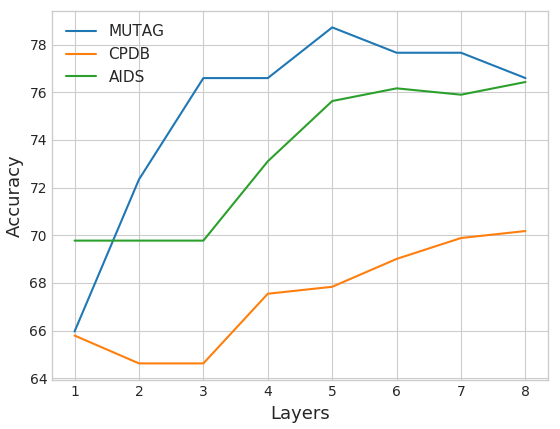}}
\caption{CGMM test accuracy as a function of the number of layers. For this experiment, we have used a CGMM without pooling and a vanilla SVM classifier trained on the concatenation of the hidden states fingerprints up to the current layer $l$. \label{fig:improvementsperlayer}}
\end{center}
\vskip -0.2in
\end{figure}

The effect of stacking on the quality of the representation learned by the CGMM can be assessed by looking at the accuracy of a linear SVM trained on the CGMM fingerprints for a varying number of layers.
 Figure \ref{fig:improvementsperlayer} depicts this information for the $3$ graph classification benchmarks, considering a CGMM model without pooling to discount its effect on model performance. 
One can clearly note that depth allows the propagation of informative structural context with a beneficial effect on accuracy on all the datasets, with an eventual performance plateau. On MUTAG, the addition of a new layer leads to a minor degradation of performance at some point, which can be explained by the greedy layer-wise and unsupervised nature of the approach. This can be easily contrasted using early stopping as in Section \ref{sect:super}.

\section{Conclusions}
\label{sect:conclusions}
We have presented a novel framework that tackles learning in the structured domain by combining generative and discriminative approaches. It works with graphs of any size and shape exploiting a rich set of concepts such as full stationarity, incremental depth of hierarchical layers and a pooling strategy. 
Importantly, CGMM does not squash the graph into a simpler representation prior to learning. The hierarchical state construction is functional to the relaxation of causal dependencies, allowing a symmetric context spreading between states representing vertexes 
and to elegantly cope with cycles. Experimental analysis shows that the model is competitive with both state-of-the-art recursive models for trees and  kernels for graphs.
The authors hope that the exploitation of the proposed framework, which can be extended in many directions, can contribute to the extensive 
use of both generative and discriminative approaches to the adaptive processing of structured data.

\section*{Acknowledgements}
D. Bacciu would like to acknowledge support from the Italian Ministry of Education, University, and Research (MIUR) under project SIR 2014 LIST-IT (grant n. RBSI14STDE).


\clearpage
\bibliography{bibliography}
\bibliographystyle{icml2018}


\end{document}